\documentclass[a4paper,11pt]{article}
\usepackage{named}
\usepackage{times}
\usepackage{graphics}
\usepackage{graphicx}
\usepackage{subfigure}
\usepackage{amsfonts}
\usepackage{amsmath}
\usepackage{amsthm,amssymb}

\newcommand{\bi}{\begin{itemize}}
\newcommand{\ei}{\end{itemize}}
\newcommand{\be}{\begin{enumerate}}
\newcommand{\ee}{\end{enumerate}}
\newcommand{\bd}{\begin{description}}
\newcommand{\ed}{\end{description}}

\newcommand{\comment}[1]{}

\newcommand{\LL}{\boldsymbol{\ell}}
\newcommand{\bx}{\mathbf{x}}
\newcommand{\bX}{\mathbf{X}}
\newcommand{\bE}{\mathbf{E}}
\newcommand{\mG}{\mathcal{G}}

\newif\ifsubmit
\submitfalse

\title{Modeling Brain Circuitry over a Wide Range of Scales}
\author{P. Fua and G. Knott\footnote{This work was supported in part the EU ERC project Micro Nano} \\ EPFL \\ CH-1015 Lausanne}
\date{}

\begin{document}
\maketitle

\begin{abstract}

If  we  are  ever to  unravel  the  mysteries  of  brain function  at  its  most
fundamental level,  we will need  a precise  understanding of how  its component
neurons connect  to each other.  Electron  Microscopes (EM) can now  provide the
nanometer  resolution   that  is  needed   to  image  synapses,   and  therefore
connections,  while Light  Microscopes  (LM) see  at  the micrometer  resolution
required to  model the 3D  structure of the  dendritic network.  Since  both the
topology and  the connection strength are  integral parts of the  brain's wiring
diagram, being able to combine these two modalities is critically important.

In fact, these microscopes now routinely produce high-resolution imagery in such
large  quantities   that  the   bottleneck  becomes  automated   processing  and
interpretation,  which is  needed for  such  data to  be exploited  to its  full
potential. In  this paper, we briefly  review the Computer Vision  techniques we
have developed at EPFL to address  this need. They include delineating dendritic
arbors from  LM imagery, segmenting  organelles from  EM, and combining  the two
into a consistent representation.

\end{abstract}

\section{Introduction}
\label{sec:intro}

\begin{figure}[t]
\begin{center} 
    \includegraphics[width=0.81\columnwidth]{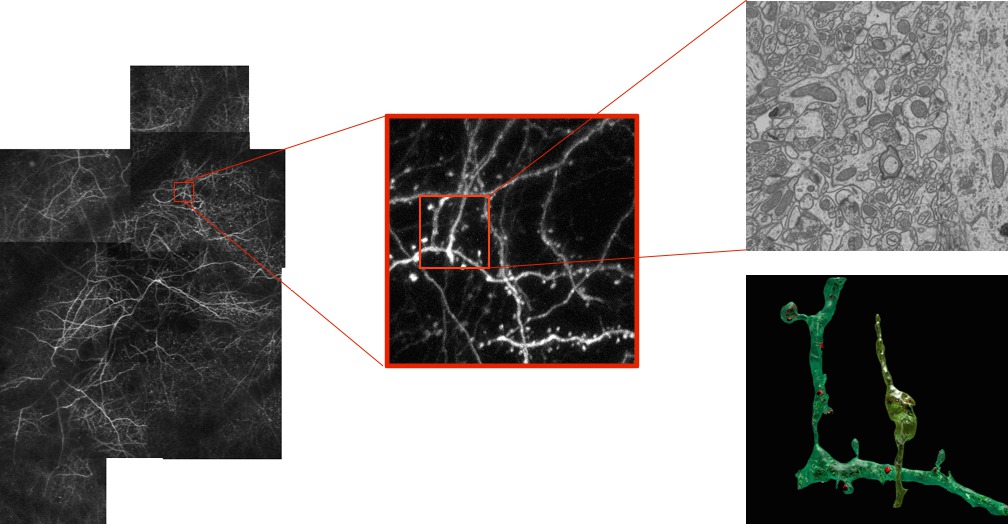} \\ 
    (a) \hspace{5cm} (b) \hspace{5cm} (c) 
\end{center} 
\vspace{-0.5cm}
\caption{Correlative Microscopy.  (a)  Fluorescent neurons in vivo  in the adult
  mouse brain imaged through a cranial window.  (b) Image stack at the $1 \mu m$
  resolution  acquired  using  a  2-photon  microscope. (c)  Image  slice  of  a
  sub-volume  at the  $5  nm$ resolution  above a  reconstruction  of a  neuron,
  dendrite, and associated organelles. }
\label{fig:correlative}
\end{figure}

As our ability to image neurons with light and electron microscopes improves, so
does our  understanding of  their form  and function. Today  we can  image large
volumes  of  both   live  and  fixed  brain  tissue  across   a  wide  range  of
resolutions. At  the micrometer  scale, light  microscopy (LM)  of fluorescently
labeled structures reveals  dendrites and axons of a subset  of neurons that can
potentially be  reconstructed revealing  their complex 3D  network, as  shown in
Fig.~\ref{fig:correlative}(b).  However, their internal structures and all their
surrounding elements remain  invisible when using this technique.   To see them,
one  must turn  to  electron  microscopes (EM).   These  provide  images at  the
nanometer scale making it possible to  visualize all the structural elements and
especially those that are important for understanding the basic connectivity and
activity  of  different  cells.    These  include  synapses,  dendritic  spines,
vesicles, and mitochondria, as depicted by Fig.~\ref{fig:correlative}(c).

These recent technologies  will therefore provide crucial  information about the
structural, functional,  and plasticity principles that  govern neural circuits.
And since  most neurological and  psychiatric disorders involve  deviations from
these principles,  such an understanding  is key to treating  them. Furthermore,
neural  circuits exhibit  a computational  power  that no  known technology  can
match.  A more thorough understanding of their complexities could therefore spur
development of new paradigms and  bio-inspired devices that would far outperform
existing ones.

However,  a major  bottleneck  stands in  the  way of  this  promise: These  new
microscopes can produce  terabytes upon terabytes of image data  that is so rich
and so complex that humans cannot analyze them effectively in their entirety. In
this paper, we will briefly present the  algorithms we have developed at EPFL to
automatically  recover the  dendritic and  axonal trees,  segment intra-neuronal
structures  from EM  images,  and  register the  resulting  models. For  further
details, we refer the interested reader to the original publications.

\section{Delineation}
\label{sec:delin}

The automated delineation of curvilinear  structures has been investigated since
the  inception  of  the  field  of  Computer Vision  in  the  1960s  and  1970s.
Nevertheless,  despite  decades of  sustained  effort,  full automation  remains
elusive when the image data is as  noisy and the structures exhibit as complex a
morphology as they  do in microscopy data. As a  result, practical systems still
require extensive manual  intervention that is both  time-consuming and tedious.
For example, in the DIADEM challenge to  map nerve cells, the results of all the
finalists  still  required   substantial  time  and  effort   to  proofread  and
correct~\cite{Diadem10,Peng11c}.

Part  of the  problem comes  from the  fact that  many existing  techniques rely
mostly  on weak  local image  evidence, and  employ greedy  heuristics that  can
easily  get trapped  in local  minima.   As a  result, they  lack robustness  to
imaging noise and  artifacts. Another common issue is  that curvilinear networks
are usually  treated as  tree-like structures without  any loops.   In practice,
however,  many interesting  networks are  not trees  since they  contain cycles.
Furthermore,  even among  those that  really are  trees, such  as neurites,  the
imaging  resolution is  often so  low that  the branches  appear to  cross, thus
introducing several spurious  cycles that can only be recognized  once the whole
structure has  been recovered.  In  fact, this is reported  as one of  the major
sources                                 of                                 error
in~\cite{Wang11b,Chothani11,Bas11,Zhao11,Turetken11,Choromanska12} and  a number
of   heuristics    have   been   proposed   to    avoid   spurious   connections
in~\cite{Chothani11,Zhao11,Turetken11}.

\subsection{Approach}
\label{sec:DelinApproach}

In  our work,  we  attempt  to overcome  these  limitations  by formulating  the
reconstruction problem as one  of solving an Integer Program (IP)  on a graph of
potential tubular paths.  As  shown in Fig.~\ref{fig:DelinMethod}, the resulting
algorithm goes through the following steps:

\begin{figure*}[t!]
  \begin{center}
    \includegraphics[width=0.81\textwidth]{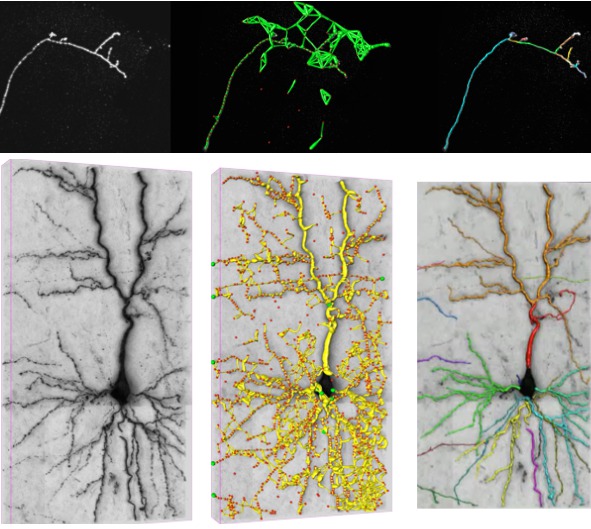}\\
    (a) \hspace{3.5cm} (b) \hspace{3.5cm} (c)
  \end{center}
    \vspace{-0.5cm}
    \caption{Delineation  in confocal  (top) and  brightfield (bottom)  imagery.
      (a) The original 3D  stacks. (b) The nodes appear as  red circles with the
      tubular paths connecting them overlaid in green and yellow.  (c) The final
      3D delineations.}
  \label{fig:DelinMethod}
\end{figure*}

\begin{itemize}
  
  \item We  first compute a  {\it tubularity} value  at each image  location and
    radius  value. It  quantifies the  likelihood  that there  exists a  tubular
    structure of this radius at that  location.  Given an 3D stack, this creates
    an 4D scale-space tubularity volume.
    
  \item We  select regularly  spaced high-tubularity points  as seed  points and
    connect pairs of them that are within a given distance from each other. This
    results   in  a   directed   tubular   graph,  such   as   those  shown   in
    Fig.~\ref{fig:DelinMethod}(b),    which    serves   as    an    overcomplete
    representation for the underlying curvilinear networks.
    
  \item  Having trained  a path  classifier using  such graphs  and ground-truth
    delineations, we assign probabilistic weights  to pairs of consecutive edges
    of a given graph at detection time.
  
 \item  We  use these  weights  and  solve an  integer  program  to compute  the
   maximum-likelihood directed subgraph of this  graph to produce a final result
   such as the one of Fig.~\ref{fig:DelinMethod}(c).
   
\end{itemize}

These  four steps  come in  roughly  the same  sequence  as those  used in  most
algorithms      that     build      trees     from      seed     points,      as
in~\cite{Fischler81b,Zhao11,Wang11b,Turetken11}, but with three key differences.
First, whereas heuristic optimization algorithms such as MST followed by pruning
or the  k-MST algorithm of~\cite{Turetken11}  offer no guarantee  of optimality,
our approach  guarantees that the  solution is within  a small tolerance  of the
global  optimum.  Second,  our  approach  to scoring  individual  paths using  a
classifier instead  of integrating pixel  values as  usually done gives  us more
robustness to  image noise and  provides peaky probability  distributions, which
helps ensure  that the global  optimum is close  to the ground  truth.  Finally,
instead  of constraining  the subgraph  to be  a tree  as many  state-of-the-art
approaches,  we  allow  it  to  contain cycles  and  instead  penalize  spurious
junctions  and   early  branch  terminations   as  described  in   more  details
in~\cite{Turetken12,Turetken13a}.

\subsection{Results}
\label{sec:DelinResults}

Here,  we demonstrate  the effectiveness  of our  approach on  the two  datasets
depicted in Fig.~\ref{fig:DelinMethod}:
\begin{itemize}
  
  \item \textit{Confocal-Axons},  8 image stacks of  Olfactory Projection Fibers
    (OPF) of  the Drosophila  fly acquired  using a  3D confocal  microscope and
    taken from the DIADEM competition.
    
  \item  \textit{Brightfield}:  6  image  stacks were  acquired  by  brightfield
    microscopy from biocytin-stained rat brains.

\end{itemize}

In both datasets, the neurites form tree structures without cycles.  However, in
the latter, disjoint  branches appear to cross, introducing false  loops, due to
the low z-resolution.  In  both cases, we used half the  stacks for training and
half for testing.  We  used a semi-automated delineation tool~\cite{Turetken13b}
to  extract  ground truth  tracings  from  the  training  stacks and  train  our
path-classifiers.  

In  Table~\ref{tab:diadem_opf},  we  compare  our  approach  (OURS)  to  several
state-of-the-art algorithms  on the confocal-axons.  They  are the pruning-based
approach  (APP2)   of~\cite{Xiao13},  the  active  contour   algorithm  (OSnake)
of~\cite{Wang11b},  the  NeuronStudio   (NS)  software  of~\cite{Wearne05},  the
focus-based  depth  estimation  method  (Focus)  of~\cite{Narayanaswamy11},  and
finally the  k-MST technique  of~\cite{Turetken11}, the last  two of  which were
finalists in  the DIADEM  competition.  For  all these  algorithms, we  used the
implementations provided by their respective authors with default parameters. We
report DIADEM  scores as  described in~\cite{Diadem10},  which were  designed to
compare  topological accuracy  of a  reconstructed tree  against a  ground truth
tree.


\begin{table}[t]
\centering
{\footnotesize
\begin{tabular}{@{}c@{\hspace{8pt}}c@{\hspace{7pt}}c@{\hspace{8pt}}c@{\hspace{7pt}}c@{\hspace{3pt}}c@{}}
\hline
\hline
& OURS & k-MST~\cite{Turetken11} & NS~\cite{Wearne05} & OSnake~\cite{Wang11b} & APP2~\cite{Xiao13} \\
\hline 
OPF4 &  \textbf{0.91} &  0.87 &  0.58 & 0.00 & 0.67 \\
\hline 
OPF6 &  \textbf{0.91} &  0.90 &  0.65 & 0.80 & 0.82 \\
\hline 
OPF7 & \textbf{0.94} &   0.91 &  0.42 & 0.68 & 0.76 \\
\hline 
OPF8 & \textbf{0.90} &   0.74 &  0.58 & 0.69 & 0.63 \\
\hline
\hline
\end{tabular}
}
\caption{\footnotesize ~DIADEM  scores on four test  stacks from
  the Confocal-Axons dataset. Each row corresponds  to an image stack denoted by
  OP$i$. Higher scores are better.}
\label{tab:diadem_opf}
\vspace{-4pt}
\end{table}

We   also   evaluated   the   APP2~\cite{Xiao13},   OSnake~\cite{Wang11b},   and
Focus~\cite{Narayanaswamy11} algorithms  on the Brightfield dataset.  Since they
do not  allow the user  to provide multiple root  vertices, the DIADEM  score of
their  output cannot  be  computed.  To  compare their  algorithms  to ours,  we
therefore used the NetMets measure of~\cite{Mayerich12} instead because it does
not rely heavily  on roots.  As the  DIADEM metric, this measure  takes as input
the reconstruction and the corresponding  ground truth tracings.  However, it is
more local because it does not account for network topology.

Table~\ref{tab:netmets_br} shows  the NetMets scores  on the test images  of the
{\it    Brightfield}     dataset.    Note     that    the     Focus    algorithm
of~\cite{Narayanaswamy11} is specifically designed  for brightfield image stacks
distorted by a point spread function.   Our approach nevertheless brings about a
systematic improvement except in one case (BRF$3$ - connectivity FPR).  However,
the algorithm  does that best in  this category does significantly  worse in the
other three.

\begin{table}[t]
\centering
{\footnotesize
\begin{tabular}{@{}c@{\hspace{0.5pt}}c@{\hspace{2.5pt}}c@{\hspace{2.5pt}}c@{\hspace{2.5pt}}c@{\hspace{3.5pt}}c@{\hspace{2.5pt}}|@{\hspace{2.5pt}}c@{\hspace{2.5pt}}c@{\hspace{2.5pt}}c@{\hspace{3.5pt}}c@{\hspace{2.5pt}}|@{\hspace{2.5pt}}c@{\hspace{2.5pt}}c@{\hspace{2.5pt}}c@{\hspace{3.5pt}}c@{\hspace{0.5pt}}@{}}
\hline
\hline
&  \vline & & BRF1 & & & &  BRF2 & &  & &  BRF3 & & \\
\hline
OURS   &  \vline & \bf{0.05}  & \bf{0.29}  & \bf{0.71} &\bf{0.65} &    \bf{0.11}  &   \bf{0.29} & \bf{0.81} &  \bf{0.78}    &    \bf{0.07}  &   \bf{0.28} & 0.77 &  \bf{0.70} \\
\hline
k-MST~\cite{Turetken11}    &  \vline  & 0.10 & 0.44 & 0.79 & 0.88 & \bf{0.11} & 0.53 & 0.84 & 0.91 & 0.13 & 0.35 & 0.81 & 0.92 \\
\hline
Focus~\cite{Narayanaswamy11} &  \vline  & 0.39 & 0.54  & 0.75 &  1.00 & 0.49 & 0.53  & 0.90 &  1.00 & 0.38 & 0.46 &  \bf{0.74} &  1.00 \\
\hline
OSnake~\cite{Wang11b} &  \vline  & 0.66 & 0.63  & 0.98 &  0.99 & 0.66 & 0.59  & 0.99 &  1.00 & 0.69 & 0.38 &  0.95 &  0.99 \\
\hline
APP2~\cite{Xiao13}    &  \vline  & 0.68 & 0.64  & 1.00  & 1.00 &  0.63 & 0.54  & 1.00  & 1.00 & 0.65 & 0.49 &  1.00 &  1.00 \\
\hline
\hline
\end{tabular}
}
\caption{\footnotesize  ~NetMets  scores  on  the  Brightfield
  dataset.  The NetMets  software outputs four values for each  trial, which are
  geometric  False Positive  Rate (FPR),  geometric False  Negative Rate  (FNR),
  connectivity FPR, and connectivity FNR, respectively from left to right. Lower
  scores are better.}
\label{tab:netmets_br}
\vspace{-10pt}
\end{table}

\section{Segmentation}
\label{sec:segm}

To observe the connectivity between  neurons electron microscopy is required. In
our work, we have used Focus Ion Beam Scanning Electron Microscopy (FIBSEM) at a
$5nm$  nearly isotropic  sampling. The  resulting image  stacks reveal  the fine
neuronal structures,  including the  synaptic contacts.  However,  segmenting EM
data poses unique  challenges in part because the volumes  are heavily cluttered
with structures  that exhibit  similar textures and  are therefore  difficult to
distinguish based solely on local image statistics.  In this section, we outline
our approach to segmenting both synapses and mitochondria. They are described in
more details in~\cite{Becker13a,Lucchi15}.

\subsection{Synapses}
\label{sec:synapse}

\subsubsection{Approach}
\label{sec:SynapseApproach}

\begin{figure*}[t]
  \begin{center}
    \includegraphics[width=0.99\linewidth]{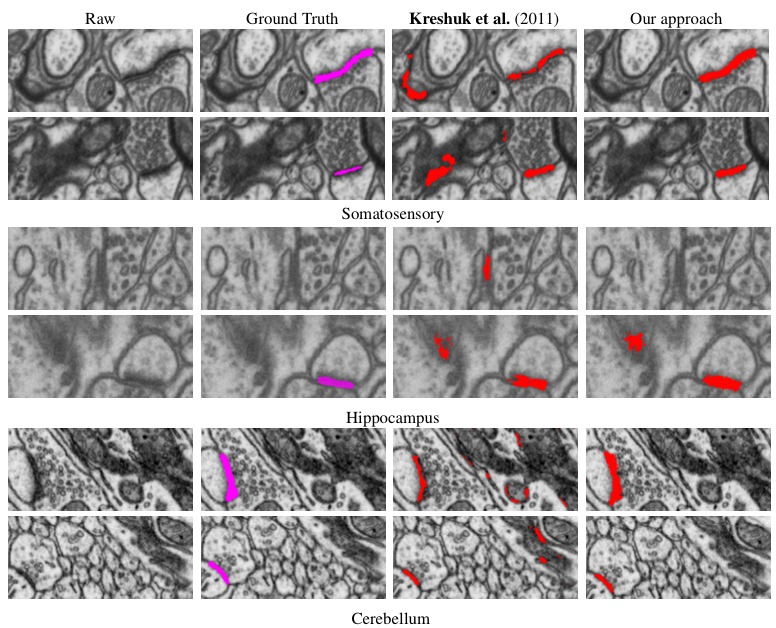}
 \end{center}
 \vspace{-0.6cm}
 \caption{Synapse  segmentations  overlaid  on   individual  slices  from  three
   different datasets  after thresholding.  Note  that our approach  yields more
   accurate results  than the  competing method  with almost  no false
   positives.}
 \label{fig:SynapseResults}
\end{figure*}

Synapses  are difficult  to distinguish  from other  structures based  solely on
local texture, as shown in Fig.~\ref{fig:SynapseResults}.  Human experts confirm
their presence by  looking for nearby for post-synaptic  densities and vesicles.
This protocol  cannot be emulated  simply by  measuring filter responses  at the
target voxel as in~\cite{Kreshuk11}, pooling features into a global histogram as
in~\cite{Lucchi11b,Narasimha09}  or  relying  on hand-determined  locations  for
feature extraction as in~\cite{Jurrus10,Venkataraju09}.

\begin{figure*}[t!] \centering
  \includegraphics[width=0.8\linewidth]{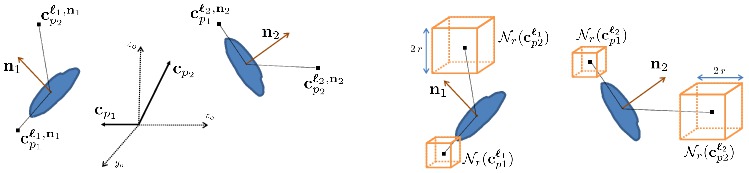}\\
  \hspace{-1.4cm} \small (a) \hspace{8cm} \small (b)
 \vspace{-0.5cm}
 \caption{Context features. (a) Relative context cue locations $\mathbf{c}_p$ in
   the  global coordinate  system  $x_o,y_o,z_o$ are  rotated  according to  the
   orientation  estimate  of  the  voxel of  interest  $\mathbf{n}_i$  to  yield
   locations $\mathbf{c}_p^{\LL_i}$ that  are consistent.  (b) At  each of these
   locations, image  channels are summed over  cubes of radius $r$  around their
   center.   Our approach  employs AdaBoost  to select  the most  discriminative
   features for synapse segmentation.}
 \label{fig:featurelocation}
\end{figure*}

To  emulate this  human  ability,  we designed  features  we  call {\it  context
  features}, which can be extracted in  any cube contained within a large volume
centered  on the  voxel  to be  classified  at 3D  location  $\LL_i$ with  local
orientation  $\mathbf{n}_i$, as  depicted in  Fig.~\ref{fig:featurelocation}(b).
They are computed in several image  channels using a number of Gaussian kernels.
This yields  more than $100,000$ potential  features and we rely  on AdaBoost to
select the most discriminative ones.

\begin{figure*}[t!]
 \centering \includegraphics[width=1\linewidth]{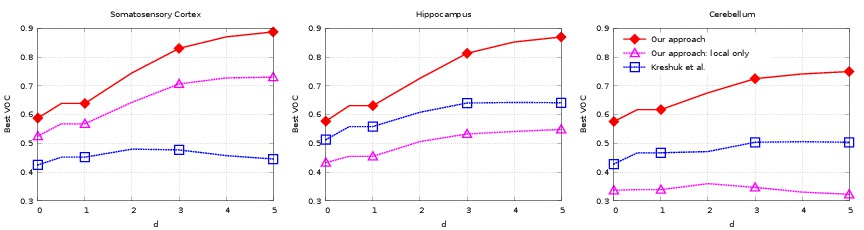}
 \vspace{-0.8cm}
 \caption{Jaccard index (VOC score) as a function of exclusion zone size $d$ for
   the  different datasets.  Our approach  outperforms the competing method for  all
   values of $d$.}
 \label{fig:SynapseVocs}
\end{figure*}

\subsubsection{Results}
\label{sec:SynapseResults}

We evaluated  our method on  three different  EM stacks acquired  from different
regions of the  adult rat brain, the Somatosensory Cortex,  the Hippocampus, and
the   Cerebellum.    Example   slices   from   each   dataset   are   shown   in
Fig.~\ref{fig:SynapseResults} along with our results.

To  evaluate   the  performance  of  our   approach  and  compare  it   to  that
of~\cite{Kreshuk11},  we  performed  a voxel-wise  evaluation  against  manually
acquired ground-truth data.  To discount  the influence of boundary voxels whose
classification may be ambiguous, we defined  a testing exclusion zone around the
labeled border of the synapse within a  distance of $d$.  The voxels within that
exclusion  zone are  ignored  and, in  Fig.~\ref{fig:SynapseVocs},  we plot  the
Jaccard index between  the ground-truth labeling and the one  the two algorithms
produce as a function  of $d$. To highlight the importance  of using context, we
plot a third curve that correspond to  our approach using only boxes centered on
the voxel to be classified, which is much worse than the other two.

\subsection{Mitochondria}
\label{sec:mitochondria}

Mitochondria  participate  in a  wide  range  of  cellular functions  and  their
morphology    and    localization    play     a    key    role    in    cellular
physiology~\cite{Scorrano10}.   Furthermore,  localization   and  morphology  of
mitochondria have  been tightly  linked to  neural functionality.   For example,
pre- and  post-synaptic presence of mitochondria  is known to have  an important
role in synaptic function, as shown in ~\cite{Lee07}, and mounting evidence also
indicates   a    close   link   between   mitochondrial    function   and   many
neuro-degenerative diseases~\cite{Knott08a,Poole08}.

New approaches  to detecting mitochondria in  EM images have therefore  begun to
appear.   For example,  in~\cite{Vitaladevuni08} a  Gentle-Boost classifier  was
trained  to  detect them  based  on  textural features.   In~\cite{Narasimha09},
texton-based mitochondria classification in melanoma cells was performed using a
variety  of  classifiers  including  k-NN,   SVM,  and  Adaboost.   While  these
techniques achieve reasonable results, they incorporate only textural cues while
ignoring  shape information.   More recently,  more sophisticated  features have
been successfully used in~\cite{Sommer10,Lucchi11b,Kumar10b} in conjunction with
either  a  Random  Forest   classifier  as  in~\cite{Kreshuk11}.  The  algorithm
of~\cite{Marquez14} could be  used to impose higher-order  shape constraints but
would  be  very difficult  to  extend  to  3D  volume segmentation  because  its
computational  requirements  are  prohibitive.    Our  approach  overcomes  this
limitation and extends these earlier techniques by explicitly modeling membranes
and  exploiting  the  power  of  our   context  features  in  a  Structured  SVM
framework~\cite{Lucchi15}.

\subsubsection{Approach}
\label{sec:MitocApproach}

To reduce  the computational complexity,  our first step  of our approach  is to
over-segment  the image  stack  into  {\it supervoxels},  that  is, small  voxel
clusters with similar intensities.  We  use the algorithm of~\cite{Achanta12} to
compute them. It lets us choose their  approximate diameter, which we take to be
on the order of the known  thickness of the outer mitochondrial membranes.  This
means  that  membranes  are  typically one  supervoxel  thick.   All  subsequent
computations are  performed on supervoxels  instead of individual  voxels, which
speeds them up by several orders of magnitude. Our task is now to classify these
supervoxels as being inside the mitochondria,  part of the membrane, or outside,
as shown in Fig.~\ref{fig:mitochondria}(b).

To   this   end,  we   introduce   a   three-class  Conditional   Random   Field
(CRF)~\cite{Lafferty01}.       It     is      defined      over     a      graph
$\mathcal{G}=(\mathcal{V},\mathcal{E})$    whose   nodes    $i\in   \mathcal{V}$
correspond to  supervoxels and  whose edges $(i,j)\in\mathcal{E}$  connect nodes
$i$ and $j$ if they are adjacent in  the 3D volume. Each node is associated to a
feature vector $x_i$ computed from the image data and a label $y_i$ denoting one
of the three classes to which a supervoxel  can belong. Let $Y$ be the vector of
all $y_i$, which we will refer to  as a {\it labeling}. The most likely labeling
of a volume is then found by minimizing an objective function of the form
\begin{equation}
\label{eq:Energy}
E^{\mathbf{w}}(Y) = \sum_{i \in \mathcal{V}} {D^{\mathbf{w}}_i(y_i)} + 
\sum_{(i,j) \in \mathcal{E}} {V^{\mathbf{w}}_{ij}(y_i, y_j)},
\end{equation}
where $D_i$ is referred  to as the unary data term and  $V_{ij}$ as the pairwise
term.  The superscript denotes the dependency  of these two terms to a parameter
vector $\mathbf{w}$.

The unary data  term $D_i$ is taken  to be a kernelized function  of the context
features of  Section~\ref{sec:SynapseApproach}.  The  pairwise term is  a linear
combination  of a  spatial  regularization  term and  a  containment term.   The
spatial term is learned from data and reflects the transition cost between nodes
$i$ and  $j$ from label $y_i$  to label $y_j$.  The  containment term constrains
the membrane class to completely enclose the inside class and to be at least one
supervoxel thick,  as originally proposed in~\cite{Delong09}.   This containment
term  is  hand-defined and  does  not  depend on  any  parameters.   The set  of
parameters  $\mathbf{w}$  to be  learned  are  therefore  the weights  given  to
individual  features in  the unary  term  and the  spatial regularization  term.
These  parameters are  learned  within the  Structured  SVM framework  discussed
above,  which requires  solving an  inference  problem on  the supervoxel  graph
$\mathcal{G}$.


\subsubsection{Results}
\label{sec:MitocResults}

\begin{figure}[t]
\begin{center} 
  \includegraphics[width=0.98\columnwidth]{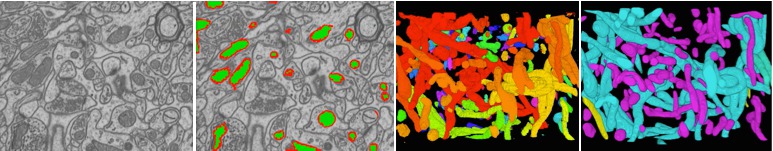} \\ 
    (a) \hspace{3.6cm} (b) \hspace{3.6cm} (c) \hspace{3.6cm} (d)
\end{center} 
\vspace{-0.5cm}
\caption{Reconstructed mitochondria.  (a)  Slice from a 3D image  stack. (b) The
  inside of  the mitochondria are  overlaid in green  and the membranes  in red.
  (c) Raw results.  (d) Edited results. The dendritic mitochondria  are shown in
  cyan and axonal ones in purple. }
\label{fig:mitochondria}
\end{figure}

Fig.~\ref{fig:mitochondria}(c) depicts the 3D reconstructions we obtained from a
$3.21  \times  m   \times  3.21  \mu  m   \times  1.08  \mu  m   $  volume.   In
Fig.~\ref{fig:mitochondria}(d),  we  show the  same  results  after having  been
proof-read and hand-corrected  by a trained neuroscientist.   The whole process,
including  generating the  training  data, took  a little  under  2 hours.   For
comparison  purposes, the  neuroscientist  re-generated  these results  entirely
manually and  that took him about  6 hours for  a similar level of  precision in
terms of  the mitochondria  volumes and  surface areas,  which are  the relevant
biological quantities.  In  other words, automation reduced  the required amount
of manual intervention by a factor 3. Going further will require deploying tools
based on deformable models  such as those of~\cite{Waneu94a,Waneu97b,Jorstad14b}
to  automatically refine  mitochondria  boundaries and  break apart  incorrectly
merged ones.

\begin{table*}[t!]
 \begin{center}
  \begin{tabular}{|@{}l@{\hspace{1mm}}|c@{\hspace{1mm}}|c@{\hspace{1mm}}|c@{\hspace{1mm}}|c@{\hspace{1mm}}|c@{\hspace{1mm}}|c@{\hspace{1mm}}|l}
    \hline
    &
    {\footnotesize Seyedhosseini'13}  &
    {\footnotesize Tsochantaridis'04} &
    {\footnotesize Wick'11} &
    {\footnotesize Lacoste'13} &
    {\footnotesize Ratliff'07} &
    {\footnotesize {\bf OURS}}
    \\
    \hline
    \hline
    {\footnotesize Hippocampus} &
    {\footnotesize 83.8\%} &
    {\footnotesize 92.7\%} &
    {\footnotesize 83.3\%} &
    {\footnotesize 92.7\%} &
    {\footnotesize 89.2\%} &
    {\footnotesize {\bf 94.8\%}} \\
\hline
    {\footnotesize Striatum} &
    {\footnotesize 83.5\%} &
    {\footnotesize 90.6\%} &
    {\footnotesize 89.6\%} &
    {\footnotesize 90.5\%} &
    {\footnotesize 88.1\%} &
    {\footnotesize {\bf 92.1\% }} \\
   \hline
  \end{tabular}\\[2mm]
 \end{center}
 \caption{\footnotesize ~Comparing  segmentation performance as measured  by the
   Jaccard  index of  the  foreground  class for  the  Striatum and  Hippocampus
   datasets against that of a number of baselines.}
 \label{tab:mitochondriaJ}
\end{table*}

To further  quantify the  performance of  our approach,  we compared  it against
other  recent  automatic  methods  on  image stacks  from  the  Hippocampus  and
Striatum,  which  are  similar  to  those   we  used  to  detect  synapses.   In
Table~\ref{tab:mitochondriaJ}, we  report the  Jaccard index for  the foreground
and membrane  class jointly, which is  representative for this task  since whole
mitochondria are  the object of  interest being segmented.   The first one  is a
very  recent mitochondria  segmentation method~\cite{Seyedhosseini13}  that does
{\it  not}  rely on  structured  learning.   Instead,  it  trains a  cascade  of
classifiers  at  different scales  and  has  been  shown to  outperform  earlier
algorithms based on Neural Networks, SVMs, and Random Forests on EM imagery. The
others correspond to different approaches to performing structured learning.  As
can be seen, we consistently outperform the competing methods.

\vspace{-0.3cm}
\section{Registration}
\label{sec:register}

Registering LM and EM stacks such as those of Fig.~\ref{fig:correlative}(b,c) is
required to identify the same region in  both images and to combine the specific
information  each modality  provides, as  discussed earlier.   However, this  is
challenging because the scale-discrepancy between the two modalities---$1000 nm$
for EM  vs $5  nm$ for  LM---produces drastic appearance  changes.  It  makes it
impractical  to use  standard registration  techniques that  rely on  maximizing
image similarity, such as those described in~\cite{Pluim03}.

Instead,  we have  proposed in~\cite{Serradell15}  a new  approach for  matching
graph structures  embedded in 3D volumes,  which can deal with  the scale-change
while being  robust to topological differences  between the two graphs  and even
changes  in  the  distances  between  vertices,  unlike  earlier  graph-matching
techniques  such  as those  of~\cite{Deng10,Smeets10}.  It  requires no  initial
position  estimate, can  handle non-linear  deformations, and  does not  rely on
local appearance or  global distance matrices.  Instead,  given graphs extracted
from the two  images or image-stacks to  be registered, we treat  graph nodes as
the features to be matched.  We model the geometric mapping from one data set to
the other as  a Gaussian Process whose predictions are  progressively refined as
more correspondences are  added.  These predictions are in turn  used to explore
the set  of all  possible correspondences  starting with  the most  likely ones,
which  allows convergence  at an  acceptable computational  cost even  though no
appearance information is available.

\vspace{-0.3cm}
\subsection{Approach}

Given  graphs  $\mG^A=(\bX^A,\bE^A)$  and $\mG^B=(\bX^B,\bE^B)$  extracted  from
image-stacks $A$ and $B$, let the $\bE$s denote edges and the $\bX$s nodes.  The
edges,  in turn,  are  represented by  dense  sets of  points  forming 3D  paths
connecting the nodes. Our goal is to  use these two graphs to find a geometrical
mapping $m$ from  $A$ to $B$ such  that $m(\bx_i^A)$ is as close  as possible to
$\bx_j^B$ in the  least-squares sense assuming that $\bx_i^A$  and $\bx_j^B$ are
corresponding voxels.

If correspondences  between points belonging  to the  two graphs were  given, we
could    directly   use    the    Gaussian   Process    Regression   (GPR)    as
in~\cite{Rasmussen06}  to  estimate  a   non-linear  mapping  that  would  yield
a~prediction of  $m$ and  its associated  variance.  In  our case,  however, the
correspondences are initially unavailable and cannot be established on the basis
of  local  image information  because  the  $A$ and  $B$  are  too different  in
appearance.  In  short,  this  means  that we  must  rely  only  on  geometrical
properties  to simultaneously  establish  the correspondences  and estimate  the
underlying non-linear transform.   Since attempting to do this  directly for all
edge points would be computationally intractable, our algorithm goes through the
following two steps:
\begin{enumerate}
 
 \item {\bf Coarse alignment:} We begin by only matching graph nodes so that the
   resulting mapping  is a  combination of  an affine  deformation and  a smooth
   non-linear  deformation.  We  initialize  the search  by  randomly picking  D
   correspondences,  which roughly  fixes  relative scale  and orientation,  and
   using them to instantiate a~Gaussian Process (GP). We then recursively refine
   it as follows:  Given some matches between $\mG^A$ and  $\mG^B$ nodes, the GP
   serves to predict where other $\mG^A$  nodes should map and restricts the set
   of potential correspondences.  Among these  possibilities, we select the most
   promising  one  and use  it  to  refine  the  GP.  Repeating  this  procedure
   recursively  until  enough  mutually  consistent  correspondences  have  been
   established and backtracking  when necessary lets us quickly  explore the set
   of potential correspondences and recover an approximate geometric mapping.

 \item {\bf Fine  alignment:} Having been learned only  from potentially distant
   graph nodes,  the above-mapping is coarse.   To refine it, we  also establish
   correspondences between  points that form  the edges connecting the  nodes in
   such a way that  distances along these edges, which we will  refer to as {\it
     geodesic}  distances, are  changed as  little as  possible between  the two
   graphs.  Because  there are many more  such points than nodes,  this would be
   extremely  expensive  to  do  from  scratch.   Therefore,  we  constrain  the
   correspondence candidates to edges between  already matched nodes and rely on
   the Hungarian algorithm of~\cite{Munkres57} to perform the optimal assignment
   quickly.

\end{enumerate}

\subsection{Results}

\begin{figure*}[t!]
  \begin{center}
    \includegraphics[width=\linewidth]{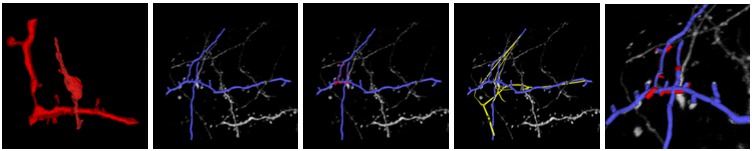}\\
    (a)\hspace{3cm}(b)\hspace{3cm}(c)\hspace{3cm}(d)\hspace{3cm}(e)
\end{center}
\vspace{-0.6cm}
\caption{ \label{fig:register} {\bf Light and electron microscopy
    neuronal trees.} {\bf (a)} Graph structure extracted from the
  electron microscopy image stack, in red. {\bf (b)} Segmented light
  microscope neurons in blue. {\bf (c)} After the non-linear
  registration process using ATS-RGM, the EM segmented neuron is
  deformed and aligned over the LM extracted neuron. {\bf (d)}
  Registration using CPD, in yellow, which falls into a~local minimum.
  {\bf (e)} A zoom over the region where the EM stack has been extracted.
  The two neurons have been completely aligned. Best viewed in color.}
\end{figure*}

Fig.~\ref{fig:register} illustrates  the two stages  of our approach  applied to
the EM and LM stacks of  Fig.~\ref{fig:correlative}.  Even though the two images
look extremely  different, our  algorithm returns  a non-rigid  deformation that
lets us correctly superpose the two stacks.  The technique is generic and allows
us  to  correctly  align  other biological  structures,  such  as  blood-vessels
networks,  that  are  non-linearly  transformed  and  extracted  with  different
techniques, without  having to pre-aligning them  and in a manageable  amount of
time.

\vspace{-0.5cm}
\section{Conclusion}
\label{sec:conc}

If  we  are  ever to  unravel  the  mysteries  of  brain function  at  its  most
fundamental level, we  will need a precise understanding of  how neurons connect
to  each other.   With  the advent  of new  high-resolution  light and  electron
microscopes, fast computers, and high-capacity  storage media, the data required
to perform this  task is now becoming available.  Electron  microscopes (EM) can
now  provide the  nanometer resolution  that is  needed to  image synapses,  and
therefore  connections,  while Light  Microscopes  (LM)  see at  the  micrometer
resolution required to  model the 3D structure of the  dendritic network.  Since
both  the arborescence  and the  connections are  integral parts  of the  wiring
diagram,  combining these  two modalities  is critically  important to  answer a
growing  need for  automated quantitative  assessment of  neuron morphology  and
connectivity.

Here,  we have  reviewed our  approach to  addressing this  daunting task.   Our
algorithms  are effective  at delineating  linear structures  in LM,  segmenting
mitochondria  and  synapses in  EM,  and  putting  the  results into  a  unified
coordinate systems  to produce  a joint  representation.  \footnote{Some  of the
  corresponding   software   can   be   downloaded  from   our   lab's   website
  http://cvlab.epfl.ch/software}  However, we  have  so far  only modeled  small
fractions  of  cells,  which  only  represent  minute  parts  of  simple  neural
circuit. Our challenge therefore is now to scale up our methods so that they can
handle  much larger  volumes, which  will involve  parallelizing them  and using
GPUs, instead of CPUs, to massively increase the processing speed.


\end{document}